\documentclass[sigconf]{acmart}

\usepackage{microtype}
\usepackage{multirow}
\usepackage{graphicx}
\usepackage{subcaption}
\usepackage{array}
\usepackage{ragged2e}

\AtBeginDocument{%
  \providecommand\BibTeX{{%
    \normalfont B\kern-0.5em{\scshape i\kern-0.25em b}\kern-0.8em\TeX}}}

\copyrightyear{2023} 
\acmYear{2023} 
\setcopyright{acmlicensed}\acmConference[CODS-COMAD 2023]{6th Joint International Conference on Data Science \& Management of Data (10th ACM IKDD CODS and 28th COMAD)}{January 4--7, 2023}{Mumbai, India}
\acmBooktitle{6th Joint International Conference on Data Science \& Management of Data (10th ACM IKDD CODS and 28th COMAD) (CODS-COMAD 2023), January 4--7, 2023, Mumbai, India}
\acmPrice{15.00}
\acmDOI{10.1145/3570991.3571065}
\acmISBN{978-1-4503-9797-1/23/01}

\begin{document}

\title{xEM: Explainable Entity Matching in Customer 360}

\author{Sukriti Jaitly}
\authornote{Work done while interning at IBM Data and AI, India}
\affiliation{%
  \institution{Carnegie Mellon University}
  \city{Pittsburgh}
  \country{USA}}
\email{sjaitly@andrew.cmu.edu}

\author{Deepa Mariam George}
\authornotemark[1]
\affiliation{%
  \institution{IBM Data and AI}
  \city{Bengaluru}
  \country{India}}
\email{deepa.george@ibm.com}

\author{Balaji Ganesan}
\affiliation{%
  \institution{IBM Research}
  \city{Bengaluru}
  \country{India}}
\email{bganesa1@in.ibm.com}

\author{Muhammad Ameen}
\affiliation{%
  \institution{IBM Data and AI}
  \city{Bengaluru}
  \country{India}}
\email{muhammed.abdul.majeed.ameen@ibm.com}

\author{Srinivas Pusapati}
\affiliation{%
  \institution{IBM Data and AI}
  \city{Bengaluru}
  \country{India}}
\email{srinivas.pusapati@in.ibm.com}

\renewcommand{\shortauthors}{Jaitly and George et al}

\begin{abstract}
Entity matching in Customer 360 is the task of determining if multiple records represent the same real world entity. Entities are typically people, organizations, locations, and events represented as attributed nodes in a graph, though they can also be represented as records in relational data. While probabilistic matching engines and artificial neural network models exist for this task, explaining entity matching has received less attention. In this demo, we present our Explainable Entity Matching (xEM) system and discuss the different AI/ML considerations that went into its implementation.
\end{abstract}



\maketitle
\section{Introduction}

\begin{figure}[!htb]
    \includegraphics[width=0.9\columnwidth]{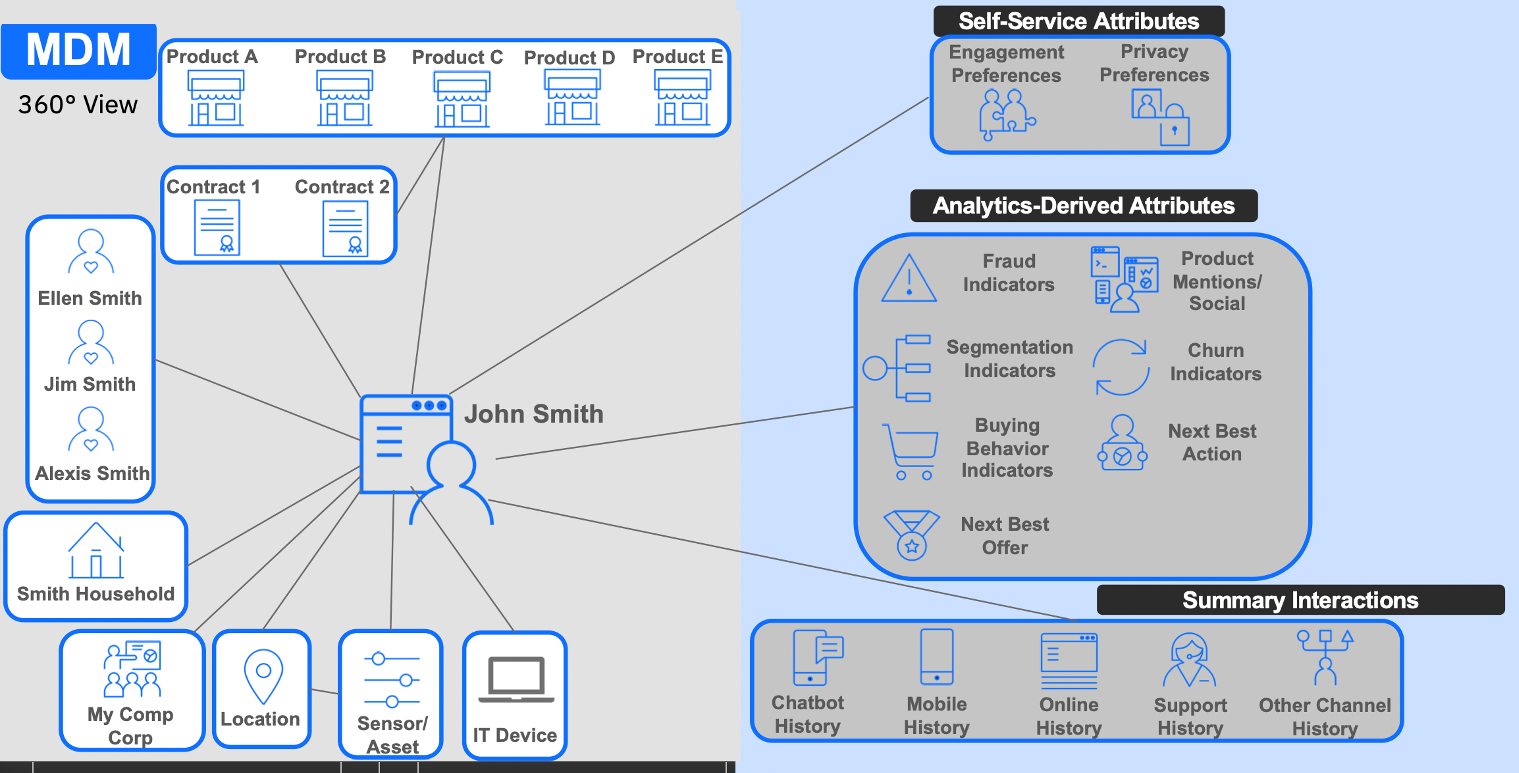}
    \caption{Customer 360 provides a 360 degree view of a customer in an enterprise data fabric}
    \label{fig:master_data_management}
\end{figure}
Entity matching is the task of predicting if two entities belong to the same real world entity. This task is critical for managing \textit{master data} in enterprises, governments and many commercial applications. Master data refers to the critical customer data that organizations maintain. Master Data Management (MDM) refers to a group of products that help organizations manage this master data. \cite{oberhofer2014beyond} introduces master data management and Customer 360 in great detail.

As shown in Figure \ref{fig:master_data_management}, customer 360 provides a 360 degree view of the customer. Entity Matching is a core component of Customer 360 with multiple use-cases as we describe in this demo. People and organization entities are of particular interest though entity matching techniques are applicable to locations, events, products, and even abstract ideas like compliance clauses and law points in legal documents.

\begin{figure*}[!htb]
\begin{subfigure}{0.48\textwidth}
    \centering
    \includegraphics[height=5cm, width=0.7\columnwidth]{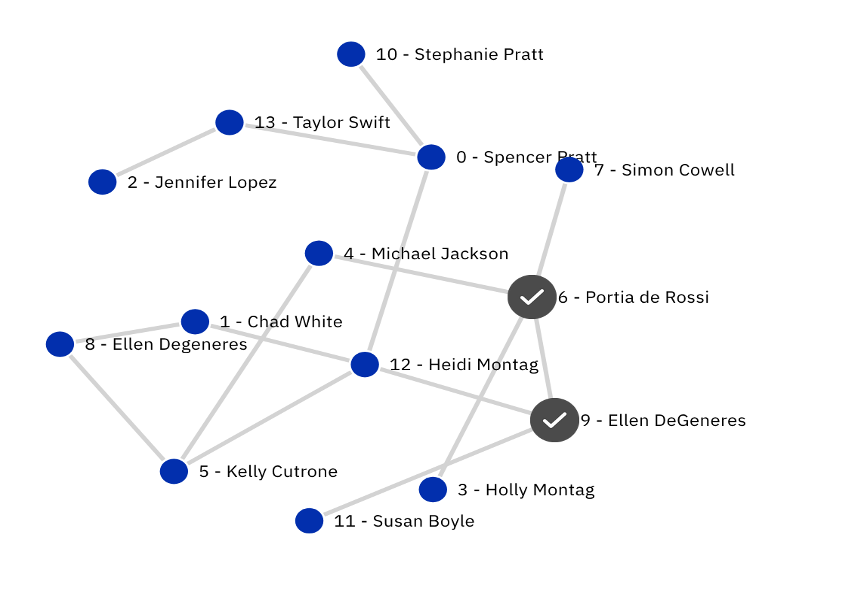}
    \caption{Graph structure around candidates to be matched}
    \label{fig:entity_matching_paths}
\end{subfigure}
\begin{subfigure}{0.48\textwidth}
    \centering
    \includegraphics[width=\columnwidth]{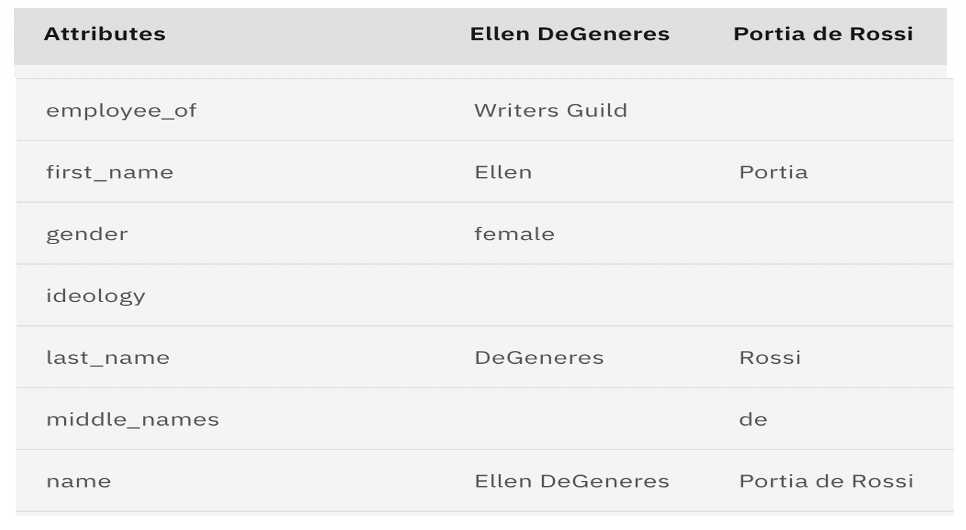}
    \caption{Comparison of node attributes}
    \label{fig:entity_matching_attributes}
\end{subfigure}
\caption{Entity Matching could be explained both by node attributes and the adjoining graph structure}
\label{currentexp}
\end{figure*}

In Customer 360, entity matching is transitive. If a record A matches with record B; and record B matches with record C;  all three records A, B and C will be linked together. This transitive linking is very useful and it helps in matching the records with partial matches. But this comes with the problem of some false positives and creating large entities of records in the system. Understanding and explaining these large entities have been a challenge so far. 

In real customer scenarios, occurrence of entities of size 1000 and above is not uncommon, though most entities have fewer records. Visualization and explanation of such large entities using path based approaches \cite{ganesan2020link} is hard because of scale. Even when the entities are not large, we need to explain why different records have been assigned to an entity.

We propose a solution to this problem by treating the relational data in a Customer 360 instance as graphs. Each record in the data becomes a node in our graph. The edges are only between records within an entity. Typically, there is an edge between a record and the representative record for the entity. Once the graph is in place, we use node embeddings from Graph Neural Networks \cite{ganesan2020link}, the scores from the probabilistic matching engine \cite{oberhofer2014beyond}, to explain entity matching.

Some of the benefits of explaining entity matching include, identifying weak links or alternatively gluing members in the entity formation, identifying false positives, identification of matching on anonymous values, and applying manual unlink rules.

But before we proceed to our solution, we'll describe our current system in Section \ref{currentsys} and two other baselines that we compared our solution against. Our demo described in Section \ref{solution}, is available from our research group page \footnote{\url{https://researcher.watson.ibm.com/researcher/view_group.php?id=11043}}.

\section{Related Work}
\cite{oberhofer2014beyond} describes the Probabilistic Matching Engine that is at the core of our entity matching solution. A number of heuristics have been developed over years, for name, addresses, phone numbers, identification numbers that are typical node attributes in Enterprise graphs. From finding edit distance, to complex statistical models, each attribute is handled differently. \cite{mudgal2018deep} presented a deep neural model for entity matching. \cite{qian2019systemer} proposed SystemER, an active learning based approach for entity resolution.

A closely related problem to entity matching in graphs is node similarity which we have described in \cite{muller2020integrated} and \cite{dhani2021similar}. Our explainability techniques using GNN models and explainers can also be applied to the node similarity task.

GNN explainability techniques in the literature include \cite{ying2019gnnexplainer}, \cite{yuan2020xgnn}, \cite{vu2020pgmexplainer},  \cite{schlichtkrull2020interpreting} and \cite{luo2020parameterized}. These techniques typically produce a subgraph as an explanation for the predicted node class or a link between two nodes.

We have in prior works attempted to substantiate GNN model predictions using information retrieval \cite{ganesan2020link}, path ranking \cite{ganesan2020explainable} and reasoner based explanations \cite{bk2021automated}. In \cite{vannur2020data}, we had used a random forest model for post-hoc explanations, while in \cite{singh2021reimagining}, we used ideas from Anchor Explanations \cite{ribeiro2018anchors} and tried on graphs models. In \cite{vanya2021automated}, we sought to automate the evaluation of these explainability techniques, since explanations are subjective and human evaluation is cost prohibitive.

\begin{figure*}[!htb]
    \includegraphics[width=\linewidth]{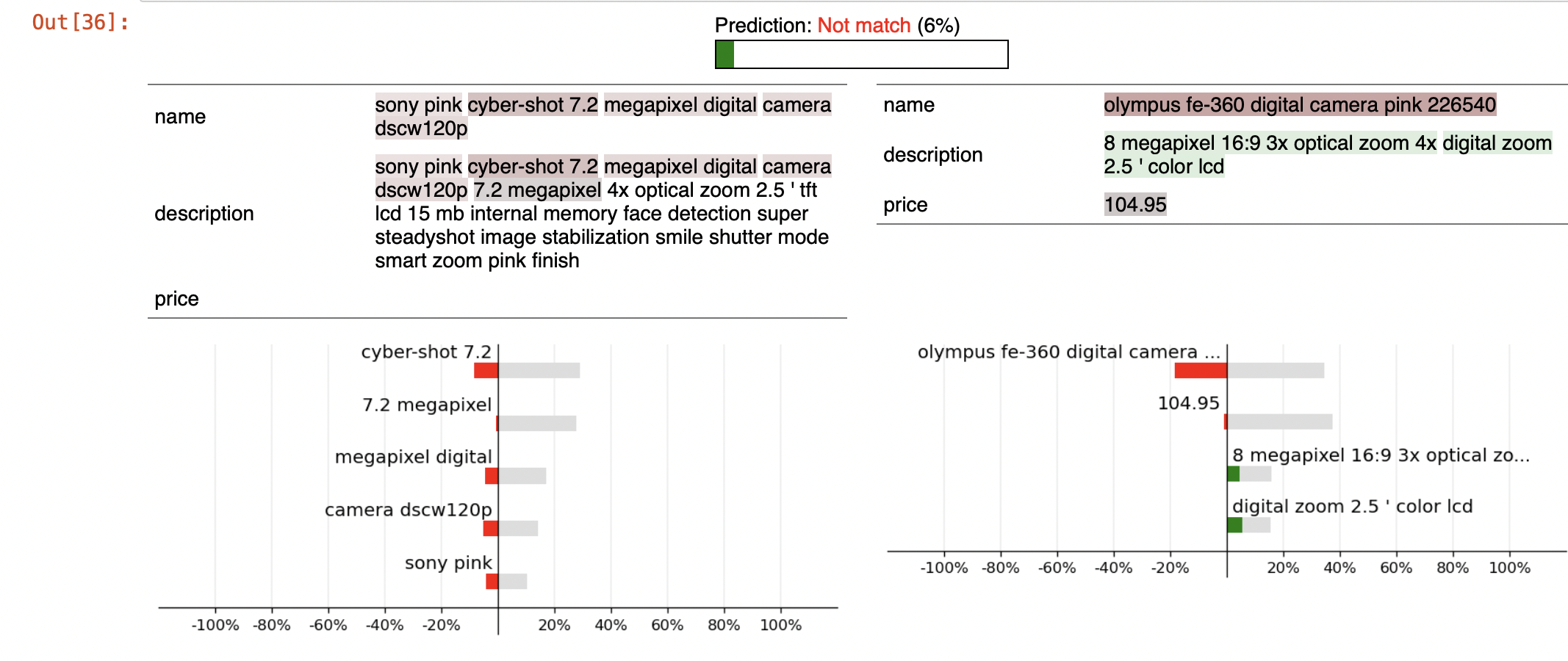}
    \caption{Explaining a non-match in Amazon-Google dataset using LEMON}
    \label{fig:lemon1}
\end{figure*}
\begin{figure*}[!htb]
    \includegraphics[width=\linewidth]{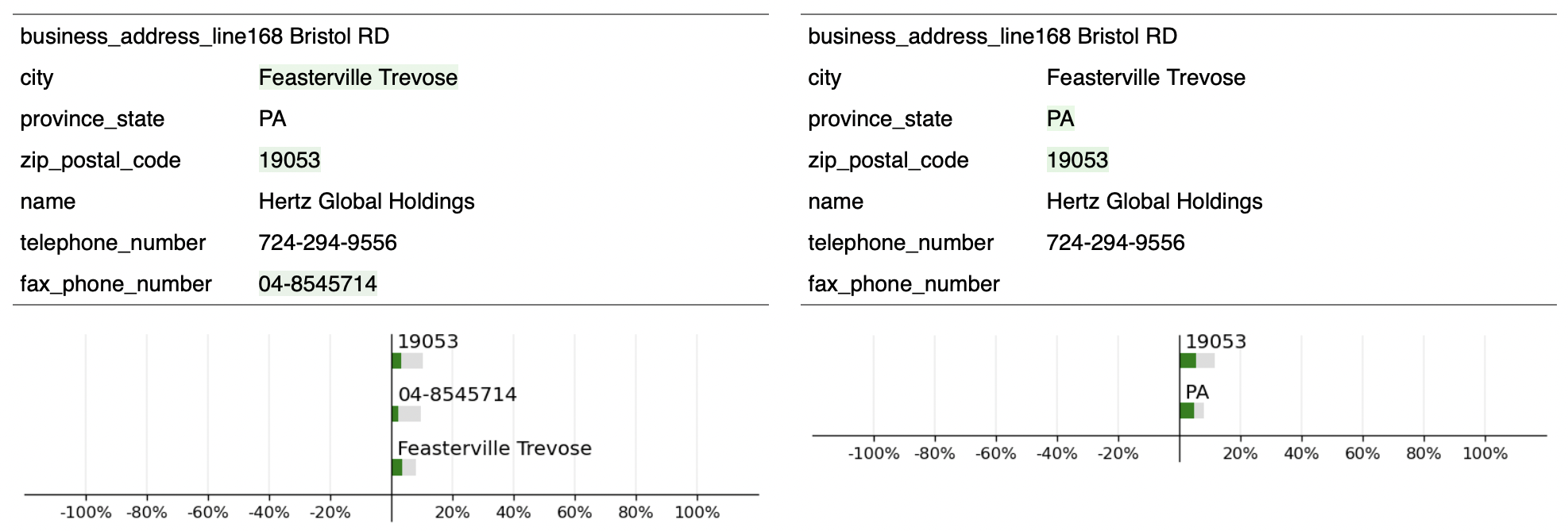}
    \caption{Explaining a match in a synthetic organization dataset using LEMON}
    \label{fig:lemon2}
\end{figure*}

\section{Baselines}
\label{baselines}

In this section, we'll first describe our current system which we seek to improve. Match 360 is an IBM product for the Customer 360 use-case and more generally for master data management.

We then implemented the models in DeepMatcher, LEMON baselines, and our own solution based on GCN and GNN Explainer. We use a combination of all three techniques to explain entity matches in different scenarios.

We evaluated the state of the art \cite{barlaug2022lemon} solution on both the Amazon-Google dataset and a synthetic organization dataset that we have created. Our synthetic dataset consists of records business name and address details with more than 10 thousand tuples.

\subsection*{IBM Match 360}
\label{currentsys}
As discussed in the previous section, transitive linking in entity matching for Master Data Management(MDM) solutions can lead to problems of false positives and large entities. Match360 is a modern MDM based solution by IBM that works with enterprise data to perform indexing, matching and linking of data from different sources (e.g. CRM, Experian, Salesforce, Web Portal), creating a 360 degree view of customer data.

Matching record pair data in Match360 requires comparing different record attributes (e.g. Name, Address, DOB, Identifier) from each pair of records to determine if they match and should subsequently be linked, based on a series of mathematically derived statistical probabilities and complex weight tables.

Such solutions that rely on Probabilistic Matching Engines(PME) for entity matching often provide very little insight into the entities making it difficult for customers to understand why such an entity was formed, or to explain them. An example of the attributes in a pair of matched records is as shown in Figure \ref{fig:entity_matching_attributes}.

\begin{figure*}[!htb]
    \includegraphics[width=\linewidth]{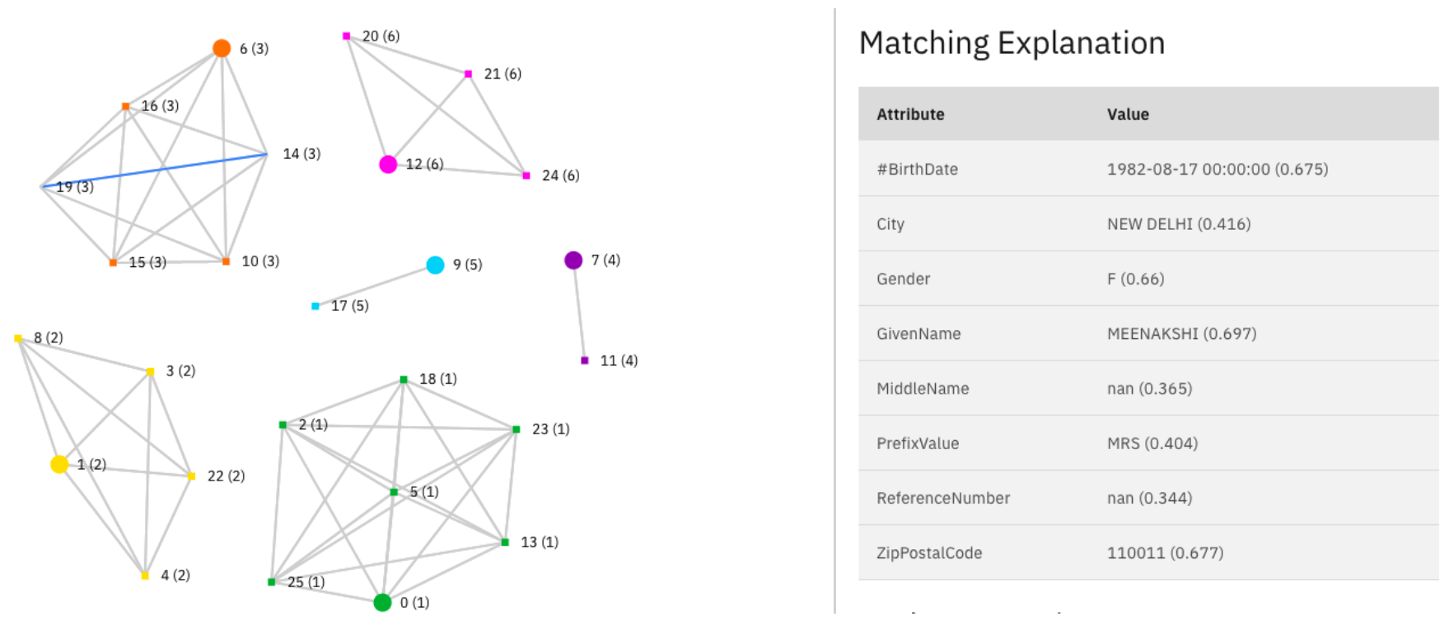}
    \caption{Explaining entities in Customer 360}
    \label{fig:large_entities}
\end{figure*}

\subsection*{DeepMatcher}

DeepMatcher \cite{mudgal2018deep} performs matching on labelled tuple pairs by training a neural network to predict matching. DeepMatcher adapts the RNN architecture to aggregate the attribute values and then compares/aligns the aggregated attribute representations. Deep Matcher trains the word embeddings using FastText.

\subsection*{LEMON}

LEMON \cite{barlaug2022lemon} is a model-independent and schema-flexible approach for evaluating explainable entity matching. This approach is effective at communicating to the user the location of the decision border, particularly in the case of non-matches.

\begin{table}[!htp]
    \begin{center}
    \begin{tabular}{lccc}
        \hline
        \textbf{Dataset}   & \textbf{Precision} & \textbf{Recall} & \textbf{F1}\\
        \hline
        \textbf{Amazon-Google} & 0.79 & 0.38 & 0.52 \\
        \textbf{Synthetic Org Dataset}  & 0.37 & 0.37 & 0.37 \\
        \hline
    \end{tabular}
    \vspace{0.5mm}
    \caption{LEMON results on entity matching datasets}
    \label{tab:entity_matching}
    \end{center}
\end{table}

As shown on Table \ref{tab:entity_matching}, we are able to roughly reproduce the performance of LEMON on the Amazon-Google dataset. But our performance on the synthetic dataset remains poor because of various challenges including the absence of any long text columns in the dataset. A typical Customer 360 dataset is similar to the examples shown in Figure \ref{fig:lemon2} and hence this is a limitation of LEMON like models from being used for this task.

Entity Matching plays a critical role in data fabric in general and data marketplace in particular. By matching records from one source to another source, we can determine relationships between datasets that may otherwise not be feasible only using metadata. We observe that LEMON explanations are particularly suited to this use-case. As shown in Figures \ref{fig:lemon1} and \ref{fig:lemon2}, we can use the explainable entity matching techniques to verify the relationship between datasets in a data fabric. We show examples of both a match and non-match. Each bar shows the positive or negative contribution of the feature to the eventual match or not-match prediction.

\section{Demo}
\label{solution}

Our solution for explaining entities in Customer 360 is to augment LEMON and other pair wise entity matching explanations with Graph Neural Network model predictions and explanations. We treat records as nodes, and edges of this graph indicate that two records match pair-wise. A cluster of all such records makes up an entity.

Our solution involves training a GNN network in batch mode on the output of the Match360, where the training would be using the pairwise comparison scores of pairs of records that PME considered during matching the dataset. We use a Graph Convolution network as implemented in \cite{you2019position} to train our model. Our entity matching system is treated as a blackbox and the trained GCN model acts a proxy for the underlying entity matching system.

After training the GNN network, during inference time, we can use the GNN model to make predictions on a limited number of pairs of nodes from the entity we intent to look into. We then proceed to explain each of these predictions using a GNN explainability solution like GNNExplainer\cite{ying2019gnnexplainer} and identify important features. These important features and their values are then passed as results to customers, helping them to examine different parts of entities and also identify any errors in the entity matching. 

We are able to explain why a record is part of this entity, by explaining why it is related to another record in the entity. By way of explanation, we highlight the important node features. Unlike a typical knowledge graph or a social network graph, there are no \textit{friend}, \textit{parent} and other kinds of relations. Hence we do not show the edge masking, but only highlight the important features using feature masking. This is in contrast to typical GNN Explanations which are sub-graphs highlighting both node and edge masks. We believe converting an explanation subgraph into a tabular form like in Figure \ref{fig:large_entities} makes it easier for end users to understand the explanation.

Simultaneously the different records that make up the nodes are displayed below the graph for any relational querying (not shown in the figure). There are multiple use-cases where this solution is being used. The default use-case is explaining why an entity has been formed by resolving multiple records (nodes) into an entity. Semantic matching where records from one source (dataset) are matched to another dataset is another use-case.

Our solution is deployed on the IBM Cloud using a Code Engine. Front end is a reactJS application while the backend is a python flask application serving REST API endpoints. Both the front end and the backend are deployed as IBM Code Engine applications. They can be bundled together onto a single container.

While these containers can be deployed both as separate microservices run on-prem, they can also be used as a cloud service on IBM Cloud. Coupled with Watson Knowledge Studio, Watson ML and Openscale, our explainability solution can be used by any customers who have Customer 360 workloads. A third monolith software approach for some of the legacy use-cases is also being considered.

In \cite{bk2021automated}, we had discussed ways to evaluate a typical explainability solution using neuro-symbolic reasoning. This is because, unlike entity matching or matching between records in different datasets, entity resolution in Customer 360 is typically a clustering problem. We leave both the neuro-symbolic evaluation of explanations and graph clustering explanations for future work.

\section*{Conclusion}
In this demo, we discussed the state of the art explainability techniques for entity matching and showed how the explanations from existing literature are inadequate for the Customer 360 use-case. We then introduced our GNN based Explainable Entity Matching (xEM) system and discussed the different AI/ML considerations that went into its implementation.


\bibliographystyle{ACM-Reference-Format}
\bibliography{references}



\end{document}